\documentclass{article}

\PassOptionsToPackage{numbers, compress}{natbib}

\usepackage[final]{neurips_data_2024}

\usepackage[utf8]{inputenc} 
\usepackage[T1]{fontenc}    
\usepackage[hidelinks]{hyperref}       
\usepackage{graphicx}
\usepackage{booktabs}
\usepackage{amsmath}
\usepackage{amssymb}
\usepackage{booktabs}
\usepackage{svg}
\usepackage{comment}
\usepackage{adjustbox}
\usepackage{booktabs,tabularx}
\usepackage[inline]{enumitem}
\usepackage{cprotect}
\usepackage[noorphans,vskip=0.5ex]{quoting}
\usepackage{longtable}
\usepackage{makecell} 
\usepackage{pifont}
\usepackage{multirow}
\usepackage{wrapfig}
\usepackage{subcaption}
\usepackage{dirtree}

\usepackage[symbol]{footmisc}

\usepackage[capitalize]{cleveref}
\crefname{section}{Sec.}{Secs.}
\Crefname{section}{Section}{Sections}
\Crefname{table}{Table}{Tables}
\crefname{table}{Tab.}{Tabs.}
\crefformat{footnote}{\textsuperscript{#2#1#3}}

\newcommand{\myparagraph}[1]{\vspace{3pt}\noindent\textbf{#1}}
\newcommand{\fulldataset}{\textsc{Fit-Coach}}
\newcommand{\qevdfulldataset}{\textsc{Qevd-Fit-Coach}}
\newcommand{\shortclipdataset}{\textsc{Qevd-Fit-300K}}
\newcommand{\qevdshortclipdataset}{\textsc{Qevd-Fit-300K}}
\newcommand{\model}{\textsc{Stream-VLM}} 
\newcommand{\qevd}{\textsc{Qevd}}
\newcommand{\sota}{state-of-the-art }
\newcommand{\nextt}{\emph{<next>} }
\newcommand{\feedbackt}{\emph{<feedback>} }

\newcommand\Tstrut{\rule{0pt}{2.6ex}}
\newcommand\Bstrut{\rule[-0.9ex]{0pt}{0pt}}

\definecolor{turquoise}{cmyk}{0.65,0,0.1,0.3}
\definecolor{purple}{rgb}{0.65,0,0.65}

\usepackage{array}
\newcolumntype{F}[1]{%
    >{\raggedright\arraybackslash\hspace{0pt}}p{#1}}%
\newcolumntype{T}[1]{%
    >{\centering\arraybackslash\hspace{0pt}}p{#1}}

\makeatletter
\DeclareRobustCommand\onedot{\futurelet\@let@token\@onedot}
\def\@onedot{\ifx\@let@token.\else.\null\fi\xspace}

\makeatother

\begin{document}

\title{What to Say and When to Say it: \\ Live Fitness Coaching as a Testbed for \\ Situated Interaction}

\author{Sunny Panchal$^{1}$\footnotemark[1] \hspace{0.35cm} Apratim Bhattacharyya$^{1}$\footnotemark[1] \hspace{0.35cm} Guillaume Berger$^{1}$ \hspace{0.35cm} Antoine Mercier$^{1}$ \\ 
\textbf{Cornelius B\"ohm}$^{2}$\footnotemark[3] \hspace{0.35cm} \textbf{Florian Dietrichkeit}\footnotemark[3] \hspace{0.35cm} \textbf{Reza Pourreza}$^{1}$ \hspace{0.35cm} \textbf{Xuanlin Li}$^{3}$\footnotemark[4] \hspace{0.35cm} \textbf{Pulkit Madan}$^{1}$ \\ \textbf{Mingu Lee}$^{1}$ \hspace{0.35cm} \textbf{Mark Todorovich}$^{1}$ \hspace{0.35cm} \textbf{Ingo Bax}$^{1}$ \hspace{0.35cm} \textbf{Roland Memisevic}$^{1}$
\\
$^1$Qualcomm AI Research\footnotemark[2] \qquad $^2$Aignostics GmbH \qquad $^3$UC San Diego
}

\maketitle

\footnotetext[1]{Authors contributed equally} \footnotetext[2]{Qualcomm AI Research is an initiative of Qualcomm Technologies, Inc.} \footnotetext[3]{Work performed at TwentyBN GmbH} \footnotetext[4]{Work performed at Qualcomm AI Research}

\begin{abstract}
Vision-language models have shown impressive progress in recent years. 
However, existing models are largely limited to turn-based interactions, where each turn must be 
stepped (i.e., prompted) by the user. 
Open-ended, asynchronous interactions, where an AI model may proactively deliver timely responses or feedback based on the unfolding situation in real-time, are an open challenge.
In this work, we present the \qevd{} benchmark and dataset, which explores human-AI interaction in the challenging, yet controlled, real-world domain of fitness coaching -- a task which intrinsically requires monitoring live user activity and providing immediate feedback.
The benchmark requires vision-language models to recognize complex human actions, identify possible mistakes, and provide appropriate feedback in real-time.
Our experiments reveal the limitations of existing state-of-the-art vision-language models for such asynchronous situated interactions. 
Motivated by this, we propose a simple end-to-end streaming baseline that can respond asynchronously to human actions with appropriate feedback at the \mbox{appropriate time.}
\end{abstract}

\section{Introduction}
Datasets that combine visual information and language have greatly contributed to advancing the abilities of AI models over the past years, 
ranging from captioning \cite{FarhadiCaptioning}, to visual questions answering \cite{VQA}, to visual dialogue \cite{DBLP:conf/cvpr/DasKGSYMPB17}, and beyond. 
Particularly impressive showcases of this progress are recent models such as GPT-4o \cite{gpt4o} and Gemini \cite{team2023gemini}, which can interact with users in real-time.

\begin{figure*}[t!]
    \includegraphics[width=\linewidth]{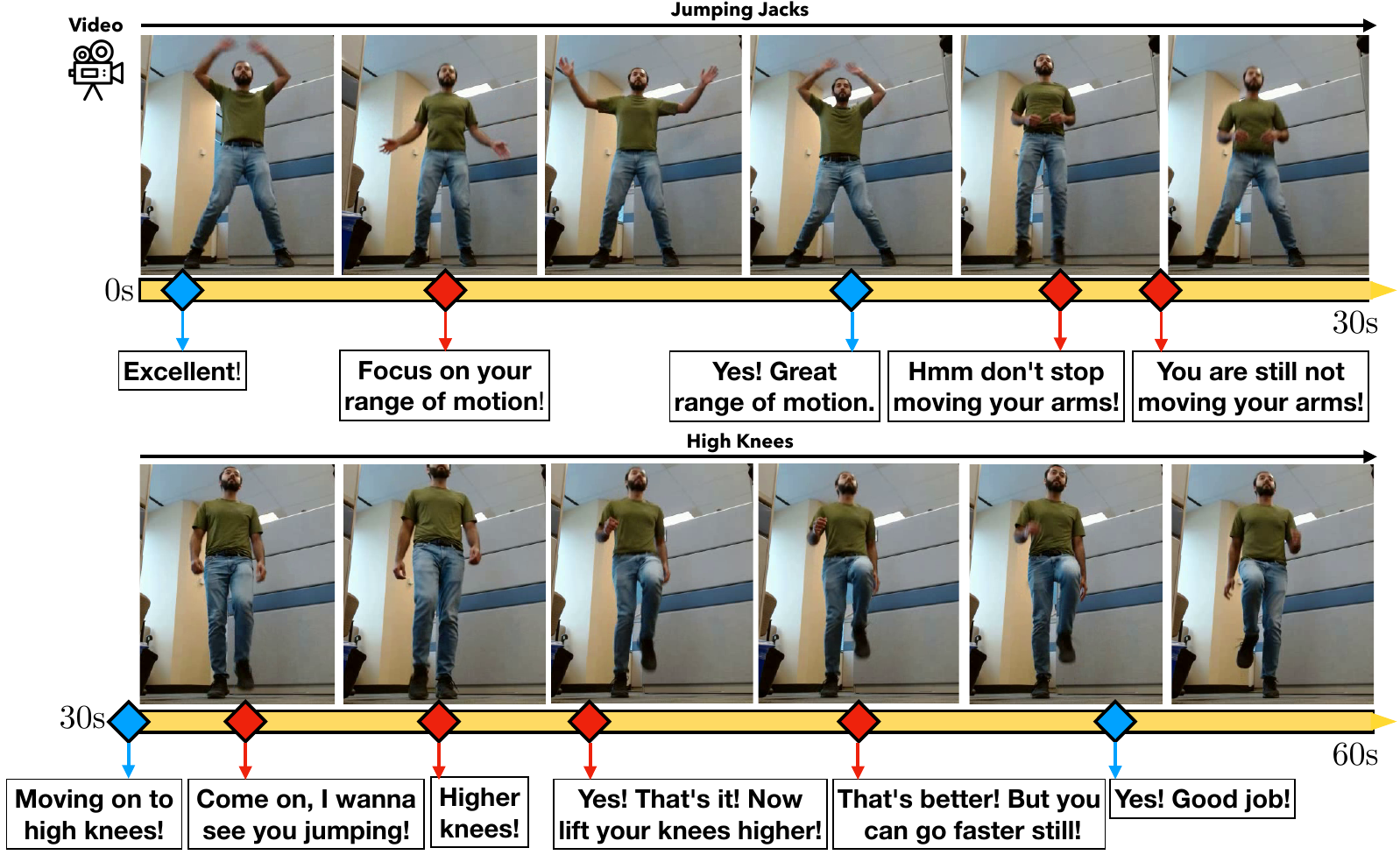}
    \setlength{\belowcaptionskip}{-0.5cm}
    \caption{Long-range interactive videos from our \qevdfulldataset{} benchmark. Live feedbacks provided to the participants are shown below each frame. Corrective feedbacks in {\color{red}red}.}
    \label{fig:finetune_dataset}
\end{figure*}

Despite the impressive recent progress, existing vision-language models still lag far behind human capabilities. 
While state-of-the-art models can be queried (e.g., through prompting) to comment on events shown in the camera stream, they lack the ability to interact asynchronously with the user as demanded by the situation, rather than only when prompted.
Such interactive scenarios that are grounded in the spatial and temporal context of an unfolding situation are commonly referred to as ``situated'' 
\cite{ammanabrolu2022dialogue,bohus2021platform,10.5555/1631171.1631258}.
Addressing such situated interactive scenarios will be a key to developing real-world assistive vision-language models.

A notable type of situated interaction is the instructional or coaching scenario, where an instructor guides a user through a complex activity, such as live fitness coaching. The real-world domain of live fitness coaching has several benefits that make it an ideal test-bed for studying situated assistive vision-language models: 
Firstly, fitness routines have a controlled structure in that users are expected to, but may not, follow a prescribed series of actions. 
Secondly, despite the structured nature of fitness routines, coaching in this domain remains a complex and unsolved problem for current vision-language systems. The nuances of human motion pose a significant challenge for these systems to effectively understand the dynamic situation and respond interactively.
Finally, live fitness coaching represents a rapidly growing real-world application. The increasing popularity of home workouts \cite{statista}, highlights the practical need and potential impact of developing effective solutions in this area. A successful vision-language model for live fitness coaching could thus offer users tangible benefits.

Currently available large-scale video datasets \cite{grauman2023ego,HeilbronEGN15,miech2019howto100m,SigurdssonVWFLG16,zhou2018towards} provide a rich set of annotations with expert demonstrations in domains such as cooking or house-hold activities.
Expert demonstrations alone are not sufficient for real-world instructional scenarios, such as live fitness coaching, where the user may make mistakes. This has been addressed recently for 
some ego-centric instructional tasks by \cite{BaoYZSBISZC23} and \cite{WangKRPCABFTFJP23}. 
Successfully guiding a user through a fitness routine additionally necessitates 
the ability to understand fine-grained human actions, to provide appropriate instructions, 
and to provide corrective feedback---grounded in those fine-grained human actions---to correct any mistakes made by the user.

Overall, our main contributions are: 
\begin{enumerate*}[label={(\arabic*)}]
    \item We propose the first large scale benchmark and dataset, Qualcomm Exercise Videos Dataset (\qevd{}), aimed at the development of video-language models for live coaching. \qevd{}\footnote{Data: \url{https://www.qualcomm.com/developer/software/qevd-dataset}; Dataset quick start guide: \url{https://github.com/varworkshop/ai_coach_fitness_2026}; and \model{} code: \url{https://github.com/Qualcomm-AI-research/FitCoach}} contains over 474+ hours of videos for fitness activity recognition and coaching. It includes short-clip videos (\qevdshortclipdataset) ($\sim$5 seconds in length) annotated with $1$M+ question-answer pairs, and long-range videos (>3 minutes in length) annotated with live feedback (\qevdfulldataset{}, cf., \cref{fig:finetune_dataset});
    \item We perform a comprehensive evaluation of state-of-the-art VLMs on the \qevdfulldataset{} benchmark, revealing that the task is largely unsolved and offers significant room for improvement; 
    \item As a step towards closing the gap towards situated interaction, we propose a novel video-language model, \model{}, which, instead of being limited to turn-based interactions, 
    can decide on-the-fly when and what to say to the user. 
    The model is trained end-to-end to perform real-time visual interaction with a user based on live camera input. 
\end{enumerate*}

\begin{table*}[t]
  \centering
  \small
  \caption{Comparison of dataset statistics. \emph{Domain}: target domain of the dataset; \emph{Human Actions}: whether the dataset contains fine grained human actions; \emph{Interactive}: whether the dataset captures interactions between two or more agents, e.g., a fitness coach and a participant; \emph{Mistakes}: whether the dataset contains correct and incorrect actions towards a task; \emph{Corrective Feedbacks}: whether the dataset contains corrective feedback provided in response to incorrect actions; \emph{Domain Expertise}: whether the dataset contains the required domain expertise to provide fine-grained feedbacks for mistakes; \emph{Length}: total length in hours.}
  \label{tab:stats}
  \begin{tabularx}{\linewidth}{Xc@{\hspace{0.2cm}}c@{\hspace{0.2cm}}c@{\hspace{0.2cm}}c@{\hspace{0.2cm}}c@{\hspace{0.2cm}}c@{\hspace{0.2cm}}c}
    \toprule
    \\
    \textbf{Dataset} & \textbf{Domain} & \raisebox{10pt}{\multirow{2}{1.2cm}{\textbf{Human Actions}}} & \textbf{Interactive} & \textbf{Mistakes} & \raisebox{10pt}{\multirow{2}{1.5cm}{\textbf{Corrective Feedbacks}}} & \raisebox{10pt}{\multirow{2}{1.3cm}{\textbf{\hphantom{t}Domain Expertise}}} & \textbf{Length} \Tstrut\Bstrut\\
    \midrule
    \multicolumn{8}{c}{{Action Recognition Datasets}} \\
    \midrule 
    NTU RGB+D \cite{ShahroudyLNW16} & Fitness & \checkmark & $\times$ & $\times$ & $\times$ &  \checkmark & --\\
    FineGym \cite{ShaoZDL20a} & Fitness & \checkmark & $\times$ & $\times$ & $\times$ & \checkmark & 708\\
    \midrule
    \multicolumn{8}{c}{{Procedural Activity Datasets}} \\
    \midrule
    YouCook2~\cite{zhou2018towards} & Cooking & $\times$ & $\times$  & $\times$ & $\times$ & $\times$ & 176\\
    Epic-Kitchens~\cite{Damen2022RESCALING} & Cooking & $\times$ & $\times$  & $\times$ & $\times$ & $\times$ & 100\\
    HowTo100M~\cite{miech2019howto100m} & Daily-life & \checkmark & $\times$ & $\times$  & $\times$ & $\times$ & 134k\\
    Ego-4D~\cite{GraumanWBCFGH0L22} & Daily-life & $\times$ & $\times$  & $\times$ & $\times$ & $\times$ & 3670 \\
    Ego-Exo4D~\cite{grauman2023ego} & Daily-life & $\times$ & $\times$ & \checkmark  & $\times$ & $\times$ & 1422 \\
    Assembly-101 \cite{SenerCSHSWY22} & Toy assm. & $\times$ & $\times$ & \checkmark  & $\times$ & $\times$ & 513\\
    \midrule
    \multicolumn{8}{c}{{Interactive AI Assistant Datasets}} \\
    \midrule
    WTAG \cite{BaoYZSBISZC23} & Cooking & $\times$ & $\checkmark$ & \checkmark & \checkmark & $\times$ & 10 \\
    HoloAssist \cite{WangKRPCABFTFJP23} & Obj. manip.  & $\times$ & $\checkmark$ & \checkmark & \checkmark & $\times$ & 166\\
    \midrule
    \qevd{} (Ours) & Fitness & $\checkmark$ & \checkmark & \checkmark & \checkmark & \checkmark & 474 \\ 
    \bottomrule
  \end{tabularx}
\end{table*}

\section{Related Work}
\myparagraph{Datasets for Activity Recognition.} 
There is a large body of work on visual activity recognition. 
This includes NTU RGB+D 120 \cite{ShahroudyLNW16}, FineGym \cite{ShaoZDL20a}, UCF101 \cite{abs-1212-0402}, Kinetics \cite{carreira2017quo}, Moments in Time \cite{MonfortVOAZRBYB20}, ActivityNet \cite{HeilbronEGN15}, AVA-Kinetics \cite{abs-2005-00214}, Charades \cite{SigurdssonVWFLG16}, Something-Something \cite{somethingsomething}, Something-Else \cite{MaterzynskaXHXW20}, and others. Unlike our \qevd{} benchmark and dataset, these datasets do not contain detailed multi-modal annotations such as questions and feedbacks, or long videos with multiple actions or events. This restricts their utility in the development of interactive video-language models.
Furthermore, while these datasets focus on a wide range of human activities, they contain only a few fitness activity related classes, if any  (\cref{tab:stats}, col 2). 

\myparagraph{Datasets for Procedural Activities.} 
The Epic-Kitchens~\cite{Damen2022RESCALING} dataset provides ego-centric videos of non-scripted daily kitchen activities, with post-hoc recorded narrations. YouCook2~\cite{zhou2018towards} provides cooking videos annotated with instructions on how to prepare specific meals, largely featuring expert chefs. HowTo100M~\cite{miech2019howto100m} provides narrated instructional videos, featuring a variety of activities. 
These datasets are not interactive (\cref{tab:stats}, col 4) as they feature first-person instructions or narrations. Furthermore, they largely feature experts and do not include mistakes likely to be made by novices (\cref{tab:stats}, col 5). In contrast, \qevd{} is interactive, features participants with diverse skill levels, and thereby, includes mistakes likely to be made by novices.

Ego-Exo4D~\cite{grauman2023ego} includes a highly diverse set of activities, performed by participants with a variety of skill levels. 
Assembly-101~\cite{SenerCSHSWY22} features videos of people with diverse skill levels assembling and disassembling 101 ``take-apart'' toy vehicles.
However, these datasets are not interactive and they do not include corrective feedbacks  (\cref{tab:stats}, col 6). \qevd{} is interactive and includes corrective feedbacks from the perceptive of the fitness coach. 

Similar to our work, 
WTAG~\cite{BaoYZSBISZC23} and HoloAssist~\cite{WangKRPCABFTFJP23} include corrective feedbacks and are focused on the development of interactive AI assistants. However, they focus on domains such as cooking or object manipulation. As they 
are recorded from an ego-centric perspective, they do not contain complex human actions (\cref{tab:stats}, col 3). Furthermore, while they include mistakes and associated corrective feedbacks, they do not include diverse examples of possible mistakes per target task (\cref{tab:stats}, col 7). \qevd{} is recorded from the perspective of a virtual fitness coach and includes fine-grained human actions -- diverse exercises and their variations including a wide diversity of mistakes per exercise.

\myparagraph{Datasets for VQA and Reasoning.}
ActivityNet Captions~\cite{HeilbronEGN15}, \textsc{Vatex}~\cite{WangWCLWW19}, TRECVID~\cite{AwadBCLFGJDSGKQ18}, HD-VILA~\cite{XueHZS00FG22}, TGIF~\cite{LiSCTGJL16}, WebVid~\cite{BainNVZ21}, Charades~\cite{SigurdssonVWFLG16}, STAR~\cite{WuYC0G21} and \textsc{Agqa}~\cite{Grunde-McLaughlin21} among others, focus largely on video captioning and question answering tasks. 
Finally, there exist a wide range of datasets on visual reasoning \cite{VQA,JohnsonHMFZG17,scienceQA}, including visual dialogue, for example \cite{DBLP:conf/cvpr/DasKGSYMPB17}, all of which, 
in contrast to our work are based on still images rather than videos.
\textsc{FixMyPose}~\cite{KimZBB21} contains annotated instructions for pose correction but is limited to pairwise images and is activity-agnostic.

\myparagraph{Models for Situated Interactions.} 
There is also a growing body of work on enabling LMs to generally reason over visual input \cite{alayrac2022flamingo,DBLP:journals/corr/abs-2211-11559,
DBLP:journals/corr/abs-2304-09842,DBLP:journals/corr/abs-2303-17580, DBLP:journals/corr/abs-2303-08128,wang2023chatvideo,zeng2023socratic,zhang2023llamaadapter,abs-2302-00923}. 
However, the existing models can answer only high-level questions about depicted scenes and objects. Models for exercise feedback are discussed in \cite{Fieraru_2021_CVPR, shao2020finegym} among others. However, such models are based on the recognition of whether or not an 
exercise was performed, or on counting repetitions, rather than providing interactive guidance and reasoning about the user's movements from a third-person viewpoint, which is the focus of this work.

\begin{table}[t!]
    \centering
    \footnotesize
    \caption{\qevd{} summary statistics. $^\dagger$The test split of the long-range videos forms our \qevdfulldataset{} benchmark. $^{\dagger\dagger}$ Average is reported per exercise for the long-range videos. $^{\dagger\dagger\dagger}$ Only a single feedback is provided at the \emph{end} of the short clips. 
    }
    \label{tab:video-dataset-details}
    \begin{tabularx}{\linewidth}{X@{\hspace{0.5cm}}c@{\hspace{0.5cm}}c@{\hspace{0.5cm}}c@{\hspace{0.5cm}}c}
    \toprule
    & \multicolumn{2}{c}{\textbf{\shortclipdataset{}}} & \multicolumn{2}{c}{\textbf{\qevdfulldataset{}}}\\
    \cmidrule{2-3}
    \cmidrule{4-5}
         &  \textbf{Train} & \textbf{Test}  & \textbf{Train} & \textbf{Test$^{\dagger}$} \\
    \midrule
     Number of Videos            &  281,660   & 16,429 & 149 & 74\\
     Unique Participants             &  1,800+     & 100  &  21 & 7\\
     Average Duration (s)       &  5.6 $\pm$ 1.1 & 5.6 $\pm$ 1.2 & 213.4 $\pm$ 3.1 & 213.7 $\pm$ 3.3\\
     Exercises per Video                &  1     & 1 & 5-6 & 5-6\\
     Total Number of Exercises          & 148  & 148 & 23 & 23 \\
     Total Classes             & 1842 & 1558 & - & - \\
     \midrule
     \multicolumn{5}{c}{Fitness Questions} \\
     \midrule
     Total High-level Questions & 535,299 & 31,326  & - & - \\
     Total Fine-grained Questions & 377,678 & 28,849 & - & - \\
     \midrule
     \multicolumn{5}{c}{Fitness Feedbacks} \\
     \midrule
     Total Feedbacks & 573,637 & 36,333 & 5,403 & 2,484 \\
     Average Feedbacks per Video$^{\dagger\dagger}$ & 2.0 $\pm$ 10.1 & 2.1 $\pm$ 10.2 & 5.0 $\pm$ 1.3 & 5.0 $\pm$ 1.2 \\
     Average Silence Period (s)$^{\dagger\dagger\dagger}$ & n/a & n/a & 5.2 $\pm$ 1.4 & 5.3 $\pm$ 1.2 \\
     Average Feedback Length (words) & 8.9 $\pm$ 5.1 & 9.2 $\pm$ 5.1  & 6.3 $\pm$ 3.8 & 6.6 $\pm$ 4.0\\
    \bottomrule
    \end{tabularx}
    \vspace{0cm}
\end{table}

\vspace{-0.1cm}
\section{Fitness Interactive Coaching Dataset and Benchmark}
\vspace{-0.1cm}
We now introduce \qevd{}, including \qevdshortclipdataset{}, and the \qevdfulldataset{} dataset and benchmark, in detail.
We begin with a detailed description of the \qevdfulldataset{} benchmark, followed by additional details of the \qevdshortclipdataset{} and \qevdfulldataset{} datasets.

\subsection{\qevdfulldataset{} Benchmark}
\label{sec:benchmark}
The \qevdfulldataset{} benchmark contains videos of participants performing a structured workout while receiving live feedback.
These feedbacks may be corrective, affirmative, or informative, depending on user activity, to improve their form and pacing as they follow the workout using the temporal structure described below.

\myparagraph{Feedback Structure.} The feedbacks in the \qevdfulldataset{} benchmark have the following structure: At the start of each exercise, acknowledging feedback is given once the user has started; otherwise, a reminder to do so is provided. A corrective feedback is provided as soon as a mistake is clearly visible. Similarly, when the user begins to correct their mistake, feedback is provided to acknowledge and guide the user to successfully correct the error. If the user is performing the exercise correctly, feedback focuses on repetition counting. When repetition counts are not possible, such as with deltoid stretches, users receive positive, encouraging feedback, regularly with an average silence period of $5$ seconds between successive feedback. Finally, at the end of each exercise, a feedback focused on the overall performance during that exercise is provided. This temporal structure ensures that feedbacks in the \qevdfulldataset{} benchmark occur at predictable time-steps, aligned to visually salient moments. The annotated feedbacks of each video were verified by a second annotator.

An example from the \qevdfulldataset{} benchmark, illustrating a trimmed workout session with two exercises—jumping jacks followed by high knees—is shown in \cref{fig:finetune_dataset}. 
The segment begins with affirmative feedback to acknowledge the participant starting the exercise. Next, a series of feedbacks to correct user mistakes and affirm their compliance are provided. Initially, the participant exhibits a low range of motion and is asked to correct this. Once the range of motion improves, the participant receives encouraging feedback. However, they then stop moving their arms, incurring another corrective feedback. Note that this feedback considers their previous arm movements. The user then moves on to the high knees exercise after being requested to do so. Here, the user receives corrective feedback to raise their knees to the appropriate height and to improve their pace. The session ends with positive encouraging feedback acknowledging that the user has currently performed the exercise. These examples highlight the highly interactive coaching sessions in our \qevdfulldataset{} benchmark and showcase the tight coupling between the participant actions and timely feedback.

\myparagraph{Statistics:} 
In total, the \qevdfulldataset{} benchmark consists of ${\sim}4.5$ hours of recorded workout sessions. Each session is ${\sim}3.5$ minutes long and consists of $5$ to $6$ randomly selected exercises arranged in $30$ second segments. 
The overall list of $23$ exercises is provided in the appendix. 
It includes a total of $7$ unique participants with a cumulative recording length of ${\sim}20$ minutes to ${\sim}1.5$ hours.

\begin{figure*}[t!]
\centering
    \includegraphics[width=\linewidth]{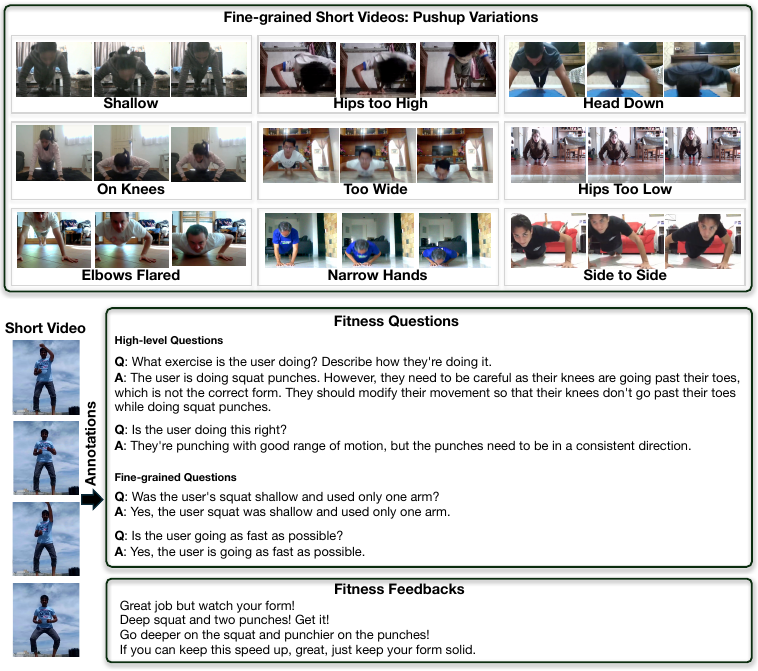}
    \setlength{\belowcaptionskip}{-0.5cm}
    \caption{Example annotations available on the short video clips from the \qevdshortclipdataset{} dataset. Annotations include question/answer pairs from our fitness questions dataset and feedback from a coaching perspective.}
\label{fig:fitness_qa}
\end{figure*}

\subsection{Fitness Datasets for Training}
\label{sec:dataset}
Together, the \qevdshortclipdataset{} and \qevdfulldataset{} datasets in \qevd{} are designed to instill domain understanding for fitness coaching and provide effective feedbacks during live coaching sessions.
They consist of three annotation types (\cref{fig:fitness_qa}), which are described in detail below.

\myparagraph{Fine-grained Fitness Activity Labels.} This includes $460+$ hours of labeled (short) videos (\mbox{\qevdshortclipdataset{}}) crowd-sourced from over $1,900$ unique participants in the wild. They cover $148$ different exercises and their variations including: varied pacing, performing common mistakes, and modified form. Exercise variations were determined top-down through consultation with expert fitness instructors. Participants were then provided detailed instructions and an accompanying reference video to perform the exercises and their pre-determined variations. Approximately\ $10$ variations were collected per exercise. We show such variations for the push-ups exercise in \cref{fig:fitness_qa} (top). Additionally, there are $49$ types of general activities such as ``grabbing a towel'' or ``drinking from a bottle''. Video lengths are in the $2$ to $10$ second range. 
There are approximately $3,500$ videos per exercise on average (${\sim}300$k clips overall) and a total of $1,800$+ fine-grained classes capturing the exercise variations and general activities. Each video and corresponding labels were manually reviewed for correctness by at least one unique crowd-worker.
These labels support training vision models for fine-grained understanding of human motion associated with fitness exercises.

\myparagraph{Fitness Questions.} In addition to the fine-grained labels, question-answer pairs querying video properties are provided for each fine-grained short video. 
This data can be used to provide further grounding of the LM's concepts in the observed visual input. 
The questions can be broadly divided into two types: high-level and fine-grained, as shown in \cref{fig:fitness_qa}. The high-level questions are directed at the overall exercise type and performance of the participant, e.g., in \cref{fig:fitness_qa}, high-level questions include ``What exercise is the user doing?'', ``Is the user doing this right?''. Fine-grained questions are designed to teach fine-grained details of exercises performed by the participant, e.g., ``Was the user's squat shallow and used only one arm?'', or ``Is the user going as fast as possible?''. These questions are generated using the Mixtral-8B-Instruct LLM \cite{abs-2401-04088} and the question generation process follows the scheme detailed in \cite{LiuLWL23a}. Overall statistics for this  subset, including a breakdown of the provided splits, can be found in \cref{tab:video-dataset-details}. 
    
\myparagraph{Fitness Feedbacks.} We provide a set of annotations from a second-person perspective to support feedback in a live coaching session. 
It contains both fine-grained short videos mentioned above (\qevdshortclipdataset{}) and additional long-range videos (\qevdfulldataset{}). For the fine-grained short videos, an average of $2$ feedbacks per video are provided (\cref{tab:video-dataset-details}). These feedbacks occur at the end of each video, as shown in \cref{fig:fitness_qa} (bottom). 
In the shown example, the feedbacks focus on improving the form of the participant, specifically, by encouraging them to squat deeper and punch with both arms. We collected an additional ${\sim}9$ hours of fitness coaching sessions following the same methodology as the \qevdfulldataset{} benchmark. These sessions contain an average of $5$ feedbacks per exercise, totaling approximately $35$ feedbacks per workout, including instructions (See \cref{tab:video-dataset-details}).

\begin{figure*}[t]
    \centering
    \includegraphics[width=0.95\linewidth]{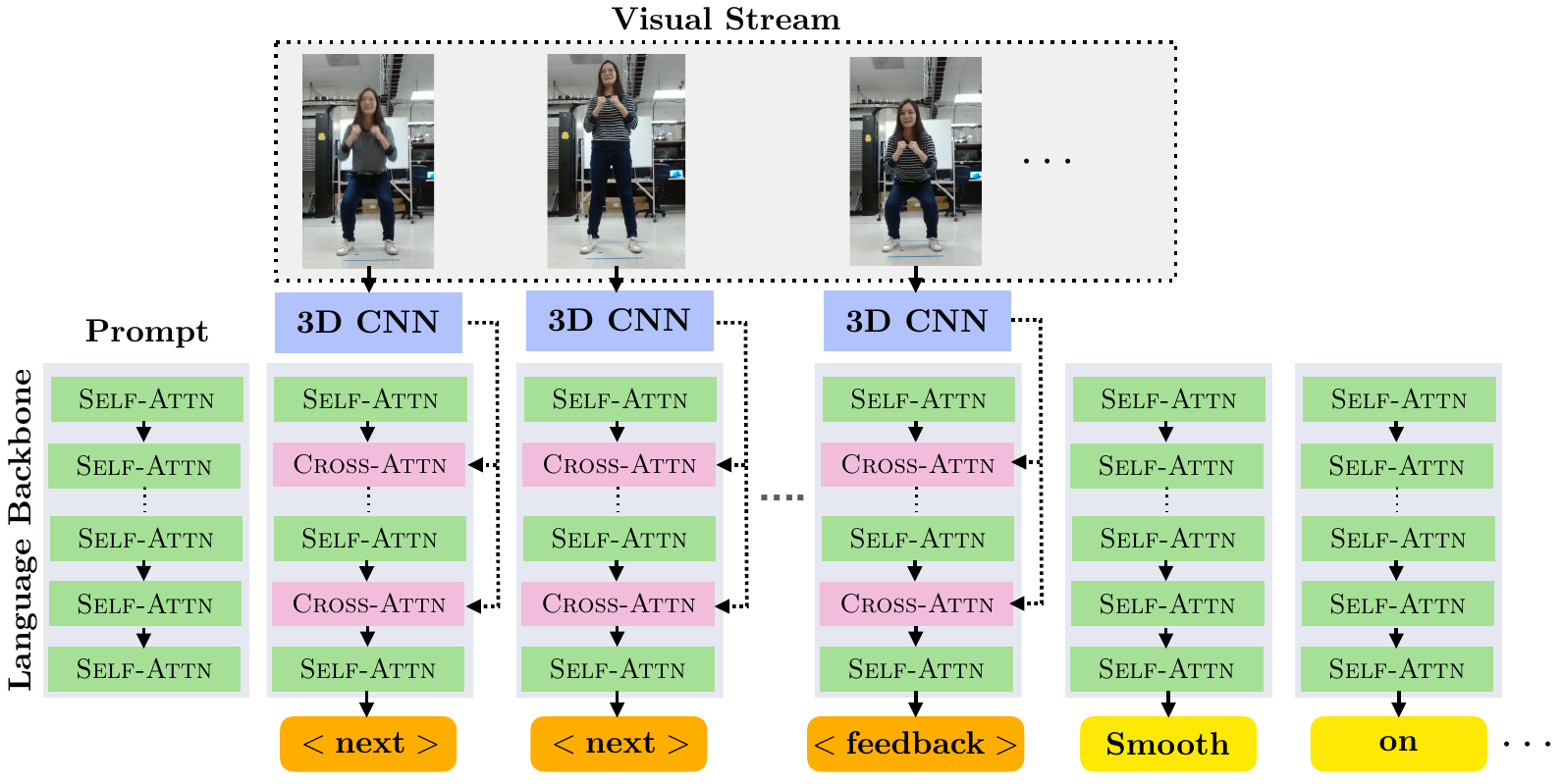}
    \caption{Architecture of the \model{} model. The visual stream is processed by a 3D CNN and the language backbone is a LLaMA-2-7B model; special action tokens (\nextt and \mbox{\feedbackt}) are highlighted in orange.}
    \label{fig:coachllama_arch}
\end{figure*}

\section{Baseline \model{}}
\label{sec:coachllama}
Current state-of-the-art vision-language models \cite{abs-2305-06500,abs-2311-17043,abs-2311-10122,abs-2306-05424,ZhangLB23,zhang2024llavanextvideo} are largely \textit{turn-based}—they take an image or video as input along with an instruction and produce a textual output.
In contrast, our \qevdfulldataset{} benchmark requires models to provide feedback proactively, i.e., without explicit prompting, based solely on the participants' actions in a \emph{streamed} setting. To this end, we propose a baseline streaming video language model, \model{}, specialized to the fitness coaching domain. It consists of a 3D-CNN-based vision backbone~\cite{sensecore, mercier2023fitness} for understanding fine-grained fitness actions and a LLaMA-2-based language backbone~\cite{touvron2023llama}. Special action tokens are introduced to enable feedback delivery without explicit prompting. We begin by describing the vision backbone in detail, followed by the special action tokens, and finally, the training scheme.

\myparagraph{Vision backbone.} Our vision backbone is designed to robustly recognize motion cues crucial for fitness coaching. This is in contrast to current state-of-the-art video-language models \cite{abs-2306-05424,ZhangLB23}, which typically use CLIP-/ViT-based vision encoders to capture scene content rather than motion. Specifically, our architecture is based on a publicly available 3D CNN \cite{sensecore, mercier2023fitness} capable of recognizing a wide range of behavior patterns, including simple exercises. 
It consists of a mix of 2D and 3D convolutional layers, ensuring that the model can pick up both motion and content information of individual frames to make predictions---both of which are relevant to provide appropriate feedbacks. Additionally, the convolutional layers are causal, making the model well-suited for the streaming setting of our \qevdfulldataset{} benchmark.
Features from the vision backbone are fused into the LM backbone through cross attention at several layers following the methodology of \cite{alayrac2022flamingo,bhattacharyya2023look}.

\myparagraph{Action tokens.} The \model{} baseline uses two special action tokens \nextt and \mbox{\feedbackt} to enable proactive feedbacks. 
The \nextt token allows the model to opt not to say anything and request the next video frame as input from the visual stream. Conversely, when the model does decide to say something, it generates a \feedbackt token. 
Through the introduction of these tokens, the model can be trained end-to-end to switch between stream observation and response generation without the need for external prompting heuristics. 
Specifically, in the coaching setting this allows the model to observe user activity and learn \textit{when} to provide feedback based on \textit{what} the user was observed doing.

In \cref{fig:coachllama_arch} the \model{} guides a user through a squats exercise. It observes the user for a few repetitions by requesting frames from the visual stream using the \nextt token. Then, at the time-step it decides to provide feedback, it outputs the \feedbackt token. This is followed by the feedback: ``Smooth on the way down $\dots$''. After the model is finished providing feedback, it requests the next video frame using the \nextt token. 

\myparagraph{Training scheme.}
\label{sec:training-scheme}
Our \model{} streaming baseline is trained end-to-end in three stages: (1) The vision backbone (3D CNN) is pre-trained on ImageNet\cite{imagenet}, followed by the \mbox{\qevdshortclipdataset{}} short-clips video collection described in \cref{sec:dataset};
(2) Next, the model is trained end-to-end on the fitness questions and fitness feedbacks annotations (excluding the long-range videos) from the \qevdfulldataset{} dataset. 
The purpose of this stage is to align the vision backbone (3D CNN) and LM with the pre-trained action recognition capability of the vision backbone. Hence, only the adapter (cross-attention layer) weights are updated; 
(3) Finally, the model is fine-tuned on long-range videos from the \qevdfulldataset{} fitness feedbacks subset with feedback annotations interleaved to reflect an interactive streaming setting. We limit the model training to $30$-second individual exercise segments and leave it to future work to train on workouts spanning multiple exercises. 
The \mbox{LLaMA-2} language backbone is fine-tuned using LoRA (dim = 32)~\cite{hu2021lora}. The 3D CNN and adapter (cross-attention layer) weights are kept frozen. Additional details are provided in the appendix. 

\begin{table*}[!t]
\small
\centering
\caption{Zero-shot evaluation on the \qevdfulldataset{} benchmark.}
\label{tab:interactive_zero_shot_eval}
\begin{tabularx}{\linewidth}{@{}Xc@{\hspace{0.5cm}}c@{\hspace{0.5cm}}c@{\hspace{0.5cm}}c}
\toprule
Method & METEOR$\uparrow$ & ROUGE-L$\uparrow$ & BERT$\uparrow$ &  LLM-Acc.$\uparrow$ \\
\midrule
InstructBLIP \cite{abs-2305-06500} & 0.047 & 0.040 & 0.839 & 1.56\\
Video-LLaVA \cite{abs-2311-10122} & 0.057 & 0.025 & 0.847 & 2.16\\
Video-ChatGPT \cite{abs-2306-05424} & 0.098 & 0.078 & 0.850 & 1.91\\
Video-LLaMA \cite{ZhangLB23} & 0.101 & 0.077 & 0.859 & 1.29\\ 
LLaMA-VID \cite{abs-2311-17043} & 0.100 & \textbf{0.079} & \textbf{0.859} & 2.20\\
LLaVA-NeXT \cite{zhang2024llavanextvideo} & \textbf{0.104} & {0.078} & 0.858 & \textbf{2.27}\\
\bottomrule
\end{tabularx}
\end{table*}

\begin{table*}[!t]
\small
\centering
\caption{Evaluation of models fine-tuned on \qevd{} on the \qevdfulldataset{} benchmark. ($^\dagger$indicates results of non-interactive models evaluated at regular intervals; $^*$indicates models fine-tuned by ourselves.)} 
\label{tab:interactive_eval}
\begin{tabularx}{\linewidth}{@{}Xc@{\hspace{0.4cm}}c@{\hspace{0.4cm}}c@{\hspace{0.4cm}}c@{\hspace{0.4cm}}c}
\toprule
Method & METEOR$\uparrow$ & ROUGE-L$\uparrow$ & BERT$\uparrow$ &  LLM-Acc.$\uparrow$ & T-F-Score$\uparrow$\\
\midrule
Socratic-LLaMA-2-7B & 0.094 & 0.071 & 0.860 & 2.17 & 0.50$^\dagger$ \\
Video-ChatGPT \cite{abs-2306-05424}$^*$ & 0.108 & 0.093 & \textbf{0.863} & 2.33 & 0.50$^\dagger$\\
LLaMA-VID \cite{abs-2311-17043}$^*$ & 0.106 & 0.090 & 0.860 & 2.30 & 0.50$^\dagger$\\
\midrule
\model{} & \textbf{0.127} & \textbf{0.112} & \textbf{0.863} & \textbf{2.45} & \textbf{0.56}\\
\midrule
\model{} (w/o 3D CNN) & 0.090 & 0.083 & 0.857 & 2.11 & 0.51\\
\model{} (w/o Pre-training) & 0.095 & 0.087 & 0.858 & 2.08 & 0.52 \\
\model{} (w/o Action-Tokens) & {0.125} & {0.110} & 0.861 & 2.41 & 0.50$^\dagger$\\
\bottomrule
\end{tabularx}
\end{table*}

\section{Experiments}
\label{sec:experiments}
In this section we evaluate current state-of-the-art (open source) video-language models and explore their limitations in the interactive streaming setting of our \qevdfulldataset{} benchmark. We also evaluate our \model{} model and highlight potential avenues to address the key challenges
associated with the \qevdfulldataset{} benchmark.

\subsection{Evaluation Metrics.}
\label{sec:eval_metrics}
The following metrics are used to capture both the fluency (``what to say'') and temporal accuracy (``when to say it'') of generated feedback.

\myparagraph{Fluency.} We use the METEOR~\cite{BanerjeeL05}, ROUGE-L~\cite{LinO04} and BERT~\cite{zhang2019bertscore} metrics to evaluate fluency. The METEOR, and ROUGE-L metrics  assess lexical similarity between the ground truth and predicted feedbacks: e.g., in the case of a corrective feedback where the person is not moving their arms, these metrics would prefer predicted feedbacks referring to the ``arm'' and ``not moving``. The BERT score on the other hand matches feedbacks at a semantic level. 

To compute these metrics, we first temporally match predicted and ground truth feedbacks. Each ground truth response is matched to the closest predicted response within a $3$ second window, maintaining their temporal order. The respective METEOR, ROUGE-L and BERT scores are then computed on only the matched feedbacks.

\myparagraph{Automated evaluation (LLM-Accuracy).} In addition to the metrics above, we employ an LLM for holistic feedback evaluation \cite{LiuIXWXZ23,abs-2306-05424}. In contrast to the metrics above, LLMs offer the advantage 
that they better correlate with human preferences~\cite{zheng2023judging}. We provide the LM with ground truth and predicted feedbacks. The LM then scores the predicted feedbacks holistically for accuracy. We use the state-of-the-art open-source LLaMA-3-70B-Instruct \cite{llama3modelcard} LLM, with scores in the range $1$ to $5$. This LLM-Accuracy metric along with METEOR, ROUGE-L and BERT metrics ensures that the predicted feedbacks match the ground-truth both at the semantic level while containing references to specific important terms. 

\myparagraph{Temporal F-Score.} To measure temporal accuracy we assess whether predicted responses occur at the correct time-step and compute a temporal F-score. Predicted responses are classified as true or false positives based on whether they temporally match ground truth responses as described for the fluency metrics. Predicted responses without a ground truth match are false positives and ground truth responses without a matching predicted response are false negatives. This allows us to calculate temporal precision, recall, and hence, the temporal F-Score. 

\begin{figure*}[!t]
    \includegraphics[width=\linewidth]{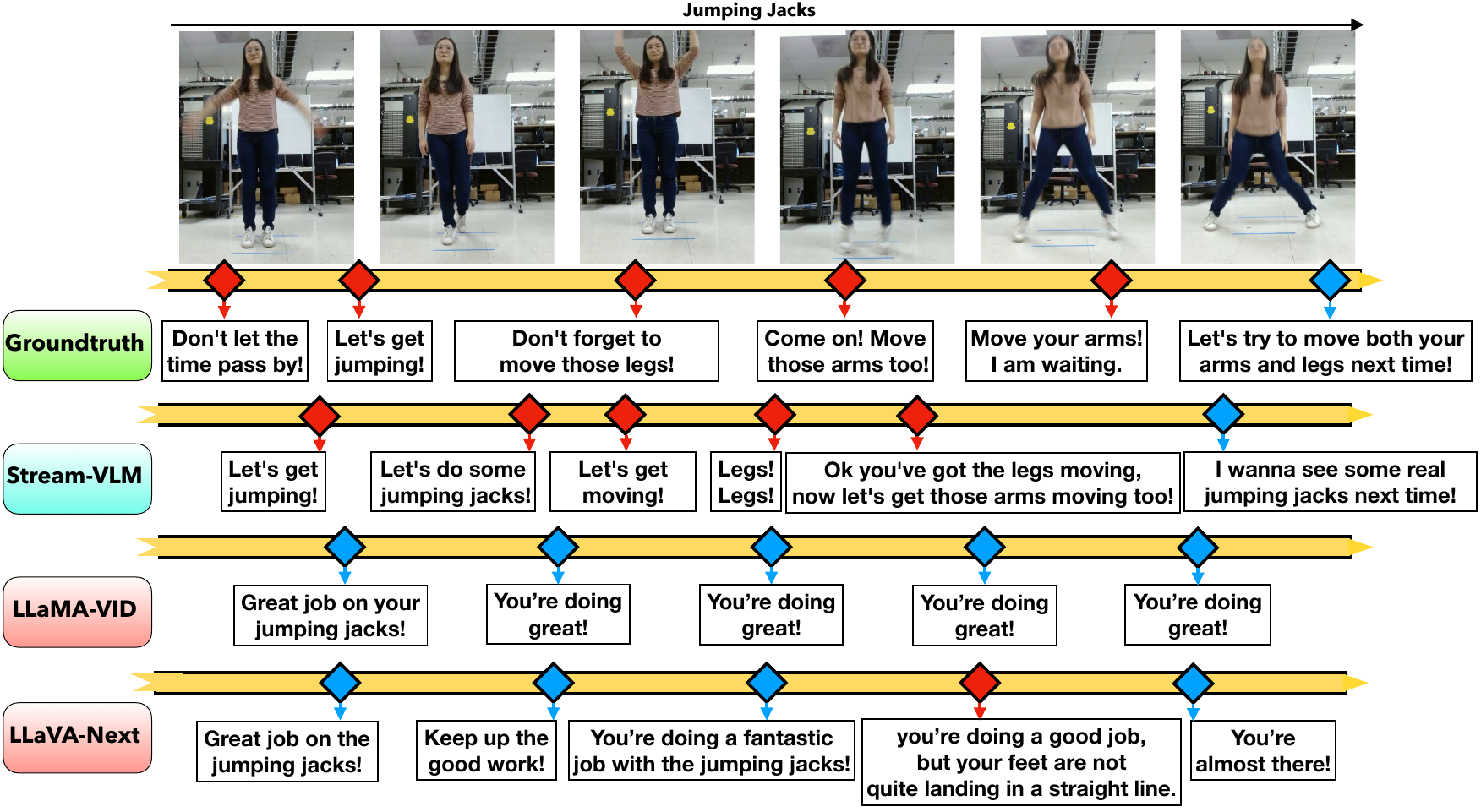}
    \vspace{-0.5cm}
    \setlength{\belowcaptionskip}{-0.3cm}
    \caption{Predicted feedbacks on the \fulldataset{} benchmark. The ``turn-based'' LLaMA-VID and LLaVA-NeXT models are unable to provide corrective feedback and instead generate overly generic and repetitive feedback. The \model{} model has learned to provide relevant feedback at the appropriate time.}
    \label{fig:finetune_eval}
\end{figure*}

\subsection{Evaluation on the \qevdfulldataset{} Benchmark}
We begin by evaluating state-of-the-art (open source) video-language models, including InstructBLIP~\cite{abs-2305-06500},
Video-LLaVA~\cite{abs-2311-10122}, Video-ChatGPT~\cite{abs-2306-05424}, Video-LLaMA~\cite{ZhangLB23}, LLaMA-VID~\cite{abs-2311-17043}, and LLaVA-NeXT\cite{liu2024llavanext},  on the \qevdfulldataset{} benchmark (see \cref{tab:interactive_zero_shot_eval}). Since these models are ``turn-based''---they cannot respond interactively to an input video---we prompt them to provide feedbacks at regular intervals. Specifically, to remain faithful to the streaming setting in our \qevdfulldataset{} benchmark, we always prompt these models (zero-shot) with the entire video, including a history of generated feedbacks, up to the latest time-step. We use an interval of $5$ seconds, equivalent to the average silence period within an exercise segment in the \qevdfulldataset{} dataset. While LLaMA-VID~\cite{abs-2311-17043} and LLaVA-NeXT~\cite{liu2024llavanext} perform best among zero-shot baselines, overall performance across all zero-shot models is weak as shown in \cref{tab:interactive_zero_shot_eval}. We present qualitative examples in \cref{fig:finetune_eval}, highlighting the repetitive and uninformative nature of the feedback provided by LLaMA-VID and LLaVA-NeXT. They are unable to provide corrective feedback at the right time largely due to their lack of fitness domain knowledge and their ``turn-based'' nature.

Next, we address the lack of fitness domain knowledge by fine-tuning Video-ChatGPT and LLaMA-VID on \qevd{} following the process discussed in \cref{sec:coachllama}. As shown in \cref{tab:interactive_eval}, fine-tuning significantly improves performance as expected. However, the performance gain is still limited by the CLIP-/ViT-based visual encoders, which are not well-suited for representing fine-grained human motion, not to mention the limitations incurred from their turn-based nature.

To deal with these issues, our \model{} baseline uses a 3D CNN trained to recognize fine-grained fitness activities and special action tokens to enable interactive feedback. We also consider the following ablations of the model:
\begin{enumerate*}[label={(\arabic*)}]
    \item instead of the 3D CNN, we use a CLIP-based encoder similar to Video-ChatGPT \cite{abs-2306-05424} (``w/o 3D CNN'');
    \item we skip pre-training the \model{} with the \qevdshortclipdataset{} short-clips fitness questions and feedbacks dataset (``w/o Pre-training''); 
    \item we use a non-interactive turn-based version without the \nextt and \feedbackt action tokens (``w/o Action Tokens'').
\end{enumerate*}
We also consider a text-only Socratic model, Socratic-LLaMA-2-7B \cite{zeng2023socratic}. In this model, we prompt the language-only LLaMA-2-7B LLM to generate feedback for the previous $5$ seconds of user activity, provided as a list of activity descriptions in $1$-second intervals. Similar to the zero-shot video-language model evaluations, the full history of described activity and generated feedbacks is included in the prompt. 
The textual description of user activity is based on the activations of the aforementioned fine-tuned 3D CNN. An example prompt is provided in the appendix.

The results in \cref{tab:interactive_eval} demonstrate that the \model{} model surpasses the performance of the other models.
Crucially, we see a significant improvement in the temporal F-score (0.59 vs 0.50) in comparison to the turn-based models. This is also illustrated in \cref{fig:finetune_eval}, where the \model{} model is shown to provide relevant corrective feedback at the appropriate time as opposed to the turn-based baselines. The improved quality of the feedbacks is also reflected in the METEOR, ROUGE-L and LLM-Accuracy metrics. The drop in both fluency and temporal accuracy resulting from the pre-training ablation (w/o Pre-training) supports the quality and utility of the fitness feedbacks and questions within our \qevdshortclipdataset{} dataset. 
Furthermore, the advantage of the 3D CNN is demonstrated by two observations: the weak performance of \model{} when the 3D CNN is ablated (w/o 3D CNN), and the strong performance of the Socratic-LLaMA-2-7B baseline. In the latter, an off-the-shelf LLaMA-2-7B model outperforms \sota vision-language models using only the activations of the 3D CNN as a prompt.

Overall, while there is significant room for improvement, these results suggest that end-to-end training is a viable path towards good performance on our \qevdfulldataset{} benchmark and more broadly, on the task of responding interactively to events within a visual stream. 

\section{Conclusion}
We propose \qevd{}, a novel interactive visual coaching benchmark and dataset, as a test-bed for real-time, real-world situated interaction, and demonstrate that this task is challenging for existing LLM-based architectures.
As a first step towards closing the gap to situated interaction, we present \model{}, a streaming vision-language model baseline that learns not only what to say, but also when to say it, based on user activity in the incoming video stream. 
Overall, we consider our work a starting point for research into end-to-end training of domain-specific interactive vision models, and hope that our data and baselines will encourage further work in this area.

\myparagraph{Privacy and ethics.} The data was collected under a direct agreement with the crowd workers, permitting research and commercial use. 
Furthermore, a detector was used on all videos to detect any issues, such as individuals in the background, followed by manual inspection of the videos that scored above a threshold, to remove such videos from the collection. 
Personal identifiable information from the videos was  removed to the extent possible, e.g., audio and meta-data.
Participants received appropriate and fair compensation for the regions where they were located.

\myparagraph{Limitations.} Our work shows that contextual, situated interactions are possible to a degree for an AI model, when a significant amount of aligned training data is made available, and the interaction is confined to a highly restricted (albeit real-world) task domain. 
    The ability to interact in broader domains, with less domain-specific training 
    data, and with higher accuracy are open research problems. 
    A related open problem is supplementing visual real-time input with speech input. 
    A further limitation is that the predictions of models trained on the data cannot be guaranteed to be free from any bias with respect to, for example, a subject's age or gender. 

\myparagraph{Broader Impact.}
    In addition to potential bias mentioned in the previous paragraph, language models can produce harmful 
    and biased content, make incorrect claims and produce wrongful advice. 
    This needs to be taken into account when interacting with, deploying or building on 
    these models, particularly in sensitive domains like fitness coaching where incorrect advice may lead to physical harm. 
    Although the grounding in visual input supports the generation of language that is 
    contextual, it is not a remedy against these deficiencies of language models. 
    It is also important to consider that any computer vision model processing visual information about human subjects could, in principle, extract information beyond what is required for the use-case, such as biometric information. 

\bibliographystyle{plainnat}
\bibliography{neurips_2024}

\clearpage
\newpage

\appendix
\section*{\centering\Large Appendix}
\section{Overview}
Here we provide: 
\begin{enumerate*}[label={(\arabic*)}]
    \item Additional qualitative examples from \qevdshortclipdataset{} and \qevdfulldataset{};
    \item Additional data collection and annotation details, including instructions provided to the crowd-workers;
    \item Additional training details of the \model{} model;
    \item Details of the prompts used for the zero-shot baselines in \cref{tab:interactive_zero_shot_eval}, including the prompt used to compute the LLM-Accuracy described in \cref{sec:eval_metrics}.
\end{enumerate*}

\section{Additional Qualitative Examples from \qevdshortclipdataset{}}
\begin{figure}[t!]
    \centering
    \includegraphics[width=\linewidth]{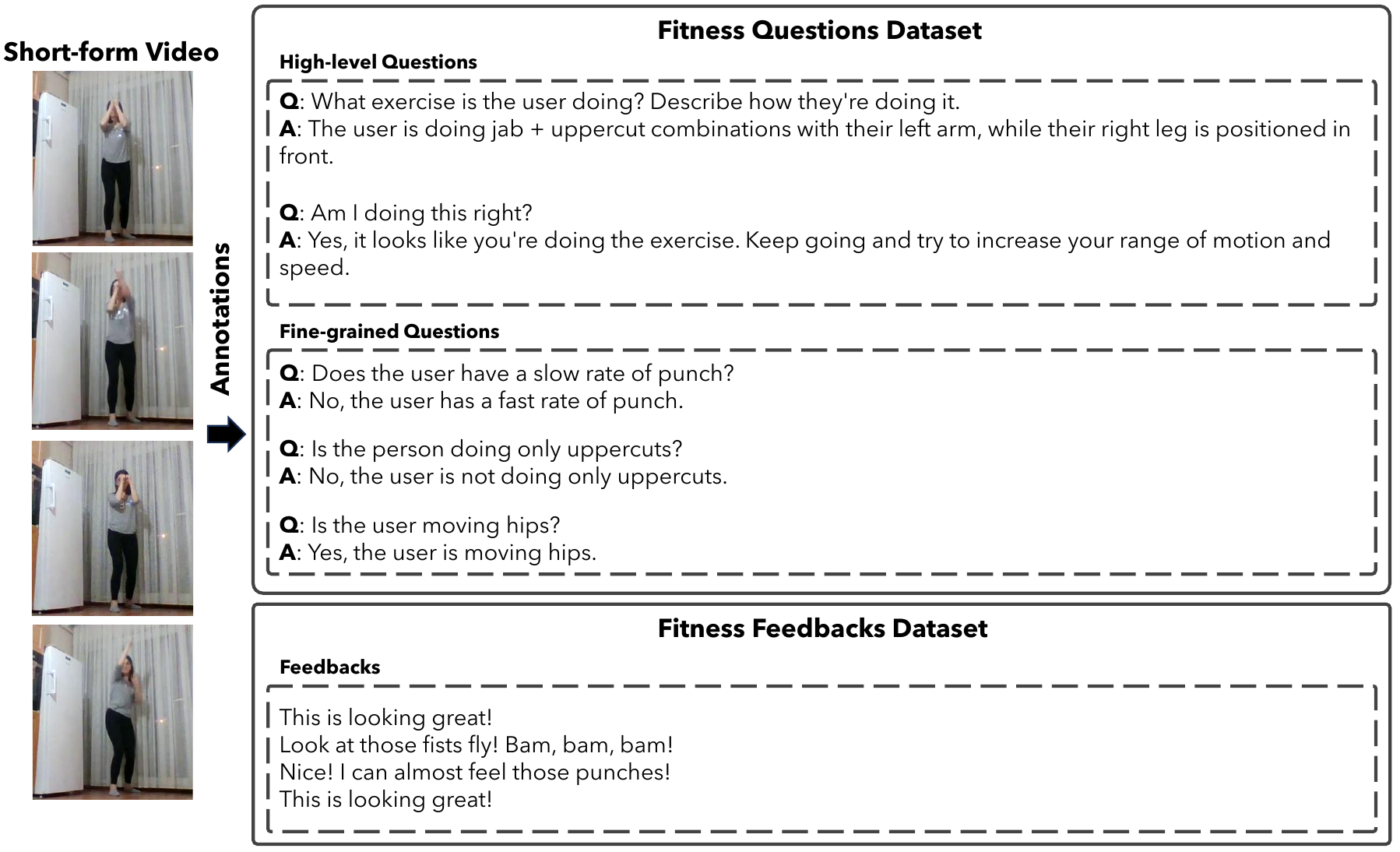}
    
    \vspace{0.25cm}
    
    \includegraphics[width=\linewidth]{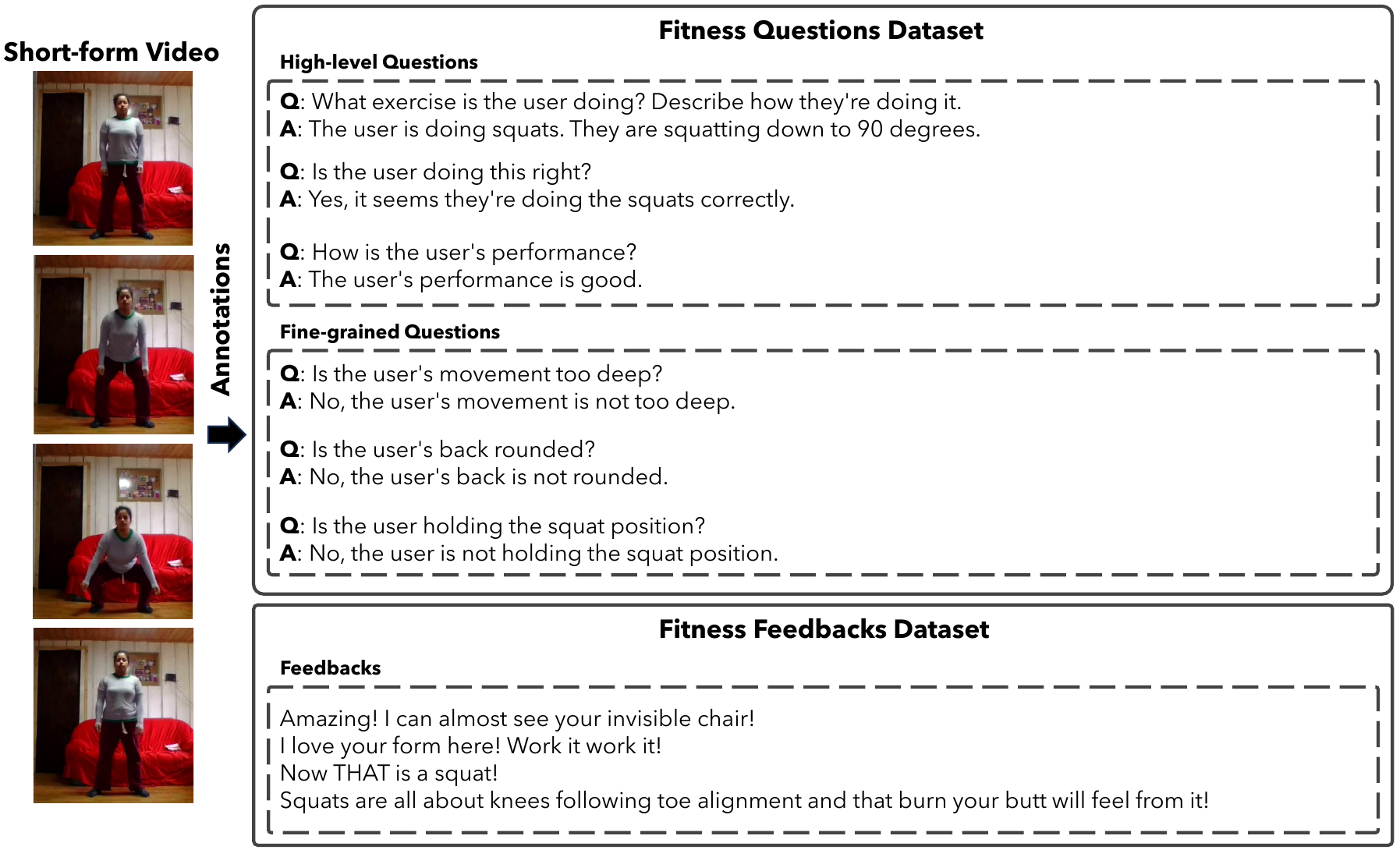}
    \caption{Additional example annotations available on the short video clips from the \qevdshortclipdataset{} dataset (see also \cref{fig:fitness_qa,fig:finetune_dataset} in the main paper).}
\label{fig:fitness_qa_supp}
\end{figure}

In \cref{fig:fitness_qa_supp}, we provide additional annotation examples from the fitness questions and fitness feedbacks on short-clip videos. As shown, high-level questions focus on overall aspects of the video, e.g., ``What exercise is the user doing? Describe how they are doing it.''. On the other hand, fine-grained questions focus on details such as the user's rate of punch (\cref{fig:fitness_qa_supp} top) or whether the user's back is rounded (\cref{fig:fitness_qa_supp} bottom). The feedbacks on these short-clips provide positive reinforcement, e.g., ``This is looking great!'' if the user is doing the exercise correctly (\cref{fig:fitness_qa_supp} top) and also provide useful advice, e.g., toe alignment while performing squats (\cref{fig:fitness_qa_supp} bottom). 

\begin{figure}[h]
\centering
\begin{subfigure}[t]{.58\textwidth}
    \centering
    \includegraphics[width=\linewidth]{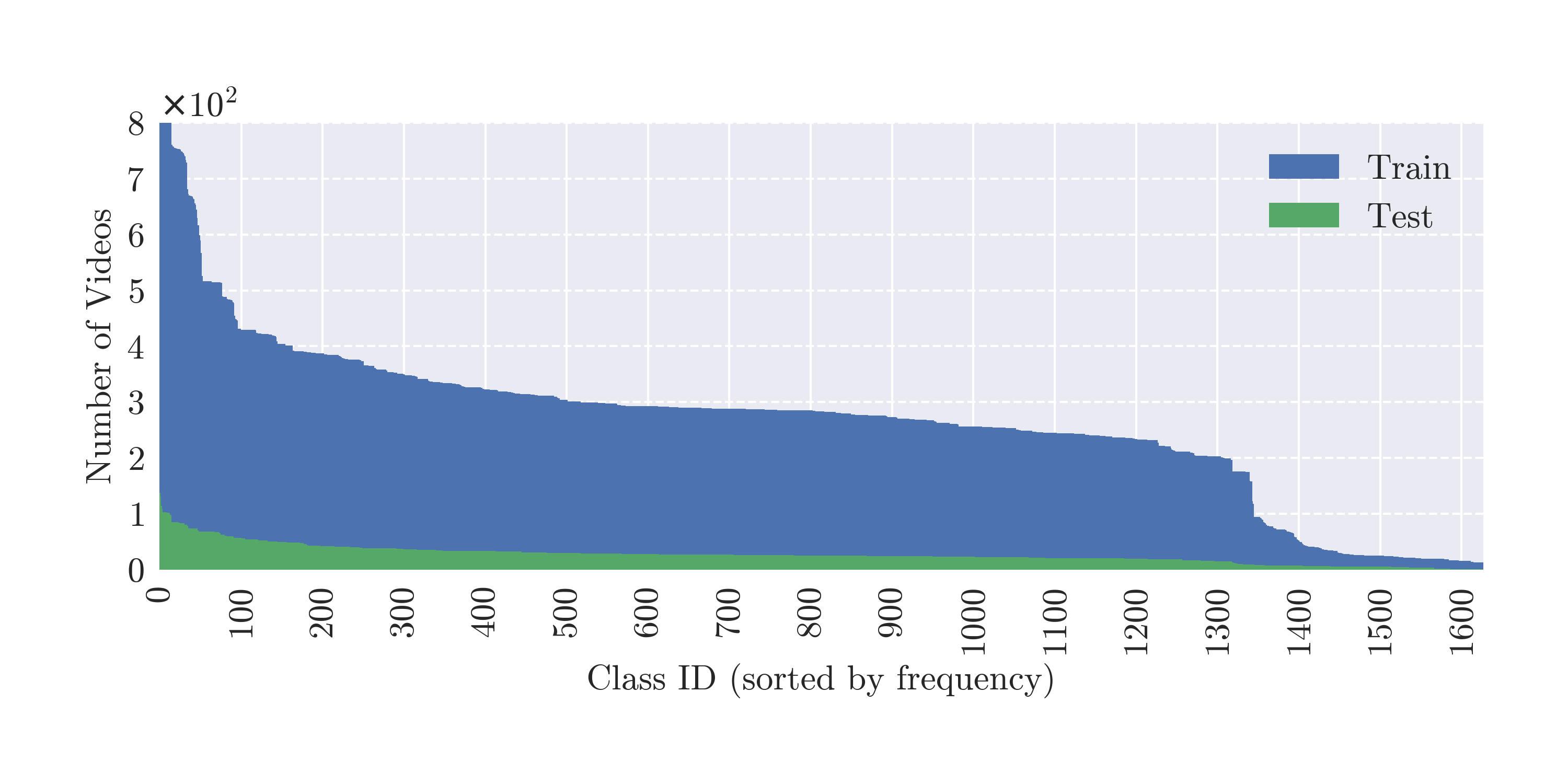}
    \caption{Distribution of fine-grained class labels.}
    \label{fig:video-class-dist}
\end{subfigure}
\begin{subfigure}[t]{.41\textwidth}
    \centering
    \includegraphics[width=\linewidth]{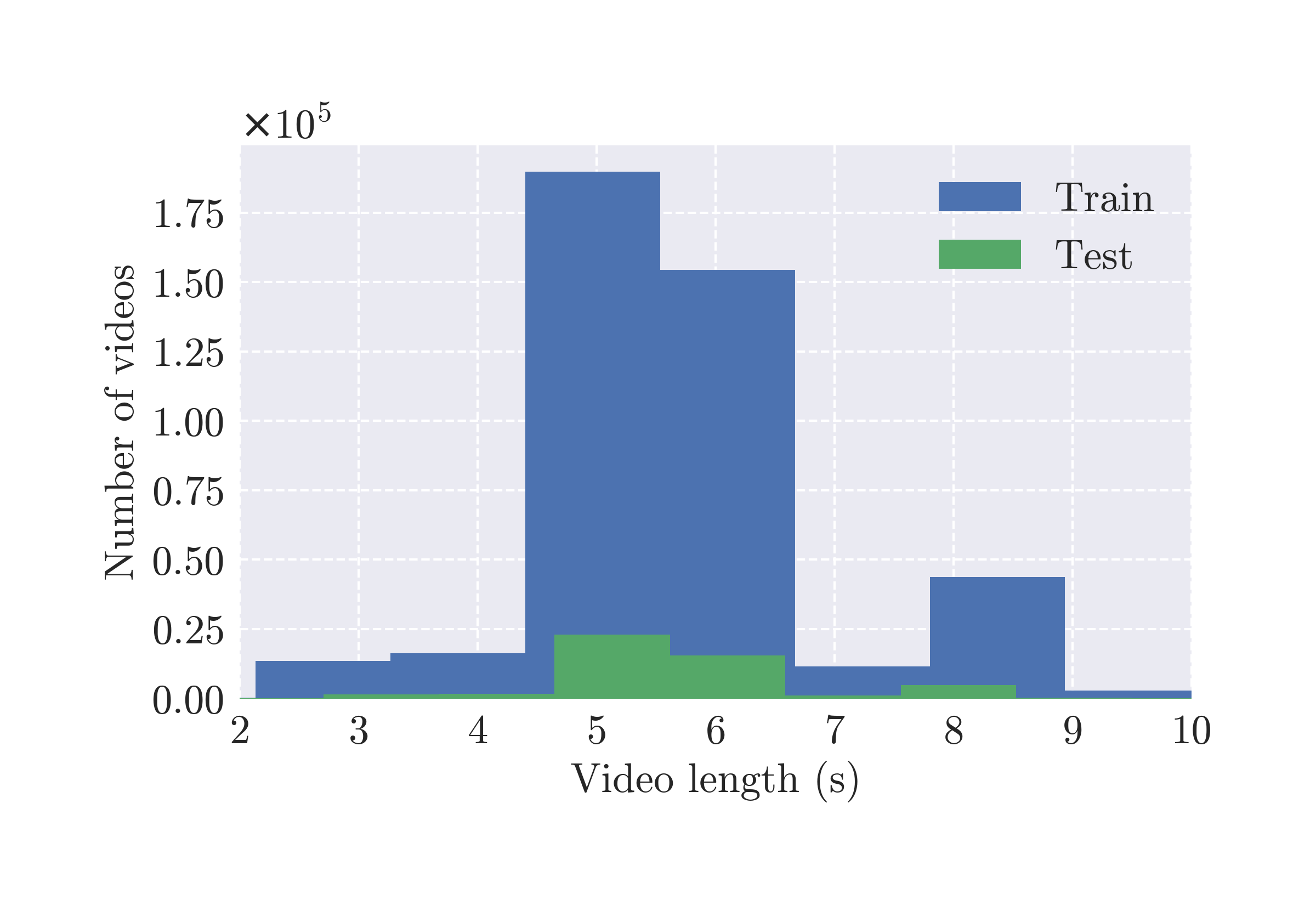}
    \caption{Distribution of clip lengths.}
    \label{fig:video-clip-dist}
\end{subfigure}
\vspace{-0.25cm}
\caption{Statistics of the short-clips in the \qevdshortclipdataset{} dataset.}
\label{fig:test}

\includegraphics[width=0.7\linewidth]{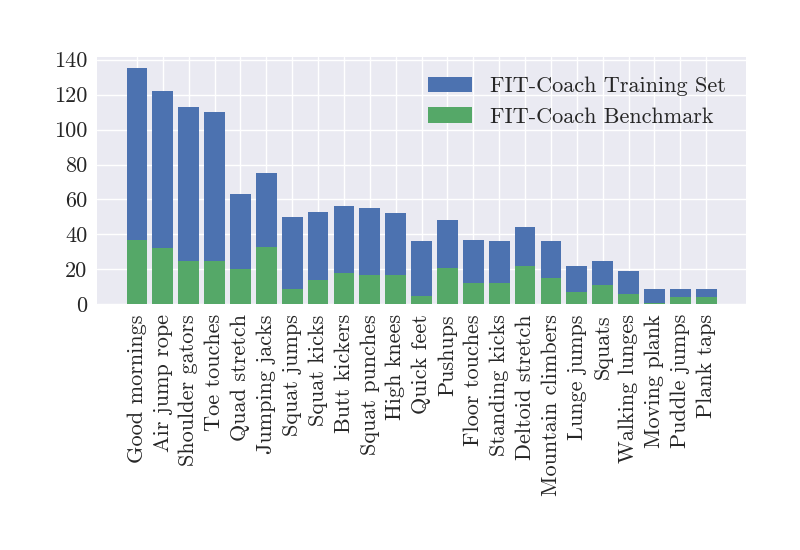}
    \vspace{-0.5cm}
    \caption{Exercise distribution in (long-form) videos from the \qevdfulldataset{} benchmark and dataset splits.}
    \label{fig:long-range-ex-dist}
\end{figure}

\begin{figure}[h]
    \centering
    \includegraphics[width=\linewidth]{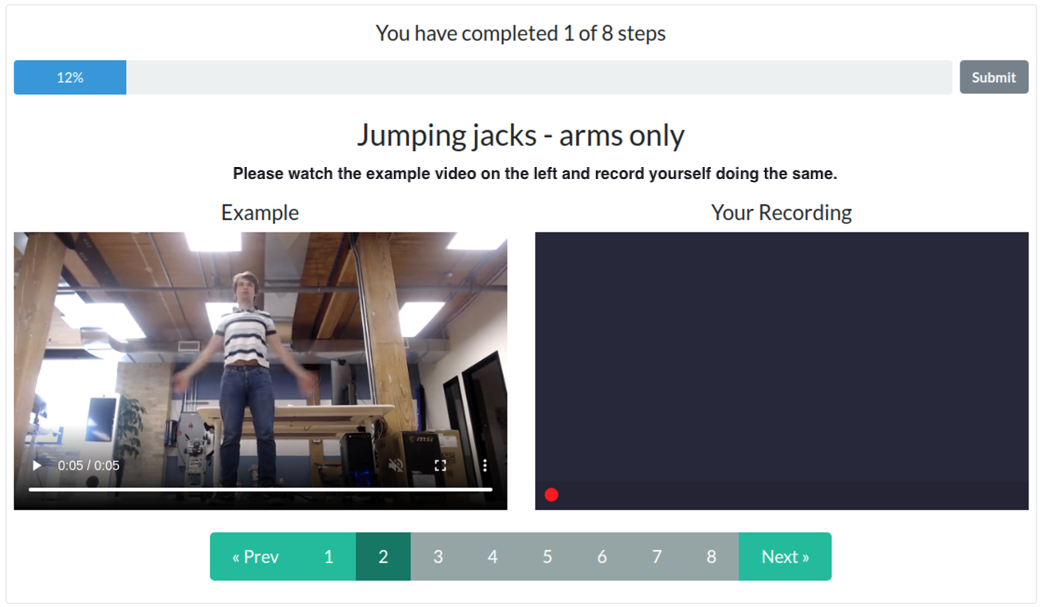}
    \caption{Simple \qevdshortclipdataset{} data collection web interface.}
    \label{fig:web-interface}
\end{figure}

\section{Additional Data Collection Details}
Additional details pertaining to the various subsets of \qevd{} are provided in this section.

\subsection{\qevdshortclipdataset{}}
\label{app:short-clips-dataset}
Here we explore the labels associated with the short video clips in the \qevdshortclipdataset{} dataset. These labels are crucial for generating fitness questions and providing feedback for short-clip videos.

\myparagraph{Video collection.} The short-clip exercise videos in the \qevdshortclipdataset{} dataset were collected through a simple web interface as shown in \cref{fig:web-interface}. The crowdworkers were shown a series of demonstration videos for the target exercise and were asked to replicate them. The exercises and their fine-grained variants were pre-determined in consultation with fitness coaches.

\myparagraph{Coarse and fine labels.} As described in \cref{sec:dataset} in the main paper, the short video clips are annotated with coarse labels and fine-grained attributes in addition to the questions and feedbacks. Labels were verified for correctness by at least one additional human annotator.
The full list of coarse labels and fine-grained attributes can be found at \url{https://developer.qualcomm.com/software/ai-datasets/qevd}.

\myparagraph{Quantitative labels.} For a subset of exercises, additional quantitative properties like speed and range of motion can be derived through simple pairwise comparative annotation campaigns. We asked crowd-workers to compare and rank randomly selected pairs of clips from $33$ exercises. The ranking allowed us to determine an ordering within an exercise. Finally, by quantizing the targets into a set of distinct groups we derived additional fine-grained labels.  Such fine-grained labels would allow downstream applications to provide more precise feedback.

\myparagraph{Feedback Labels.} Additionally, we collect feedbacks for the short videos from the perspective of a fitness coach. These feedbacks correspond to the fine-grained events for $136$ exercises, totaling over $8500$ unique feedbacks.
    
\myparagraph{High-level Fitness Questions.} As described in the main paper, the fitness questions dataset consists of multiple question-answer pairs per short-clip video (\cref{fig:fitness_qa,fig:fitness_qa_supp}) which are generated semi-automatically using the coarse-grained labels, fine-grained attributes and quantitative labels.
We convert these raw labels and attributes to conversational-style questions answer pairs using Mixtral \cite{abs-2401-04088}. An example prompt for the high-level question ``What exercise is the user doing?'':
\newpage
\begin{quote}
   \scriptsize
    Identify the exercise being performed from the provided templated exercise names and provide the answer in response to the question "What exercise is the
    user doing?".
    
    The provided names will be in one of these formats (description after `--' not part of the template):\\
    1) <exercise\_name> (<variant>) -- where variant may describe the side of the body being used or describing the position of the body.\\
    2) <exercise\_name>
    
    For example:\\
    Templated name: spider man pushup\\
    Descriptive response: The user is doing spider man pushups.\\
    Templated name: lunges (left leg out in front)\\
    Descriptive response: The user is doing lunges with their left leg in front.\\
    
    Give a descriptive response for the following:\\
    Templated name: <templated name>\\
    Don't provide an explanation.
\end{quote}

where, <templated name> is filled in with each query label.

\myparagraph{Fine-grained Fitness Questions.} The fine-grained questions per short-clip video (\cref{fig:fitness_qa,fig:fitness_qa_supp}) are generated using the annotated fine-grained attributes, quantitative labels and feedbacks.
We convert these raw labels and attributes to conversational-style questions answer pairs using Mixtral \cite{abs-2401-04088}.
    
\subsection{\qevdfulldataset{}}
The set of exercises included in the \qevdfulldataset{} benchmark and dataset (long-range videos) is shown in \cref{tab:workout-template}. The 23 unique exercises are highly diverse, of varying difficulty levels and require a wide variety of motion types by the participants. Note that this set of exercises is a subset of all exercises included in the \qevdshortclipdataset{} dataset. We show the distribution of these exercises across splits in \cref{fig:long-range-ex-dist}. 

\myparagraph{Video Collection.} 
Long-range workout videos were collected by letting users perform workout sessions consisting of three $1$-minute sections, where each section consisted of two exercises. Candidate exercises per section are shown in~\cref{tab:workout-template}. Users were instructed to perform specific exercises for the indicated duration while facing the camera to imitate a virtual fitness coaching setup.
They were also instructed to perform common mistakes and their corrections (similar to the short-clips). 
Temporally aligned textual coaching instructions were subsequently cleaned-up and verified using a simple web interface.
All feedbacks were reviewed by at least one additional annotator.

\begin{table}[hb!]
    \begin{center}
    \small
    \caption{Exercise options for the warm-up, main, and cool-down sections of the workout.}
    \vspace{0.25cm}
    \begin{tabular}{|l|l|}
        \hline
        \textbf{Section} & \textbf{Exercise Candidates}\\
        \hline
        Warm-up & jumping jacks, high knees, butt kickers, \\
        &air jump rope, good mornings\\
        \hline
        Main & push-ups, plank taps, moving plank, \\
        & squats, walking lunges, lunge jumps, \\
        &puddle jumps, mountain climbers, floor touches, \\
        &quick feet, squat jumps, squat kicks, \\
        &standing kicks, boxing squat punches\\
        \hline
        Cool-down & deltoid stretch, quad stretch, shoulder gators, \\
        & toe touchers \\
        \hline
    \end{tabular}
    \label{tab:workout-template}
    \end{center}
\end{table}

\section{Training Details of the \model{} Model}
Here, we provide additional details of training the \model{} model described in \cref{sec:training-scheme} in the main paper.

\subsection{Vision Backbone}
After pre-training the 3D CNN model on ImageNet~\cite{imagenet}, we fine-tune the 3D CNN model on the fine-grained labels available on the short-clips dataset (full list in~\cref{app:short-clips-dataset}). The 3D CNN model was trained for $500$ epochs with a batch size of $32$, a learning rate of $1\times10^{-4}$, an Adam optimizer~\cite{kingma2014adam,loshchilov2017decoupled}, and a frame-wise cross entropy loss objective. 

\myparagraph{Dataset pre-processing:} The 3D CNN vision backbone is trained on random crops of 64 frames which corresponds to roughly $4$ seconds. 
Additionally, RGB values were normalized and random color jittering was applied to each input channel. Video clips were randomly flipped horizontally with a probability of $50\%$. In the event a video was flipped, labels identifying bilateral states (such as ``left'' and ``right'') were fixed accordingly.

\subsection{Short-clip Videos}
We fine-tune the \model{} model end-to-end after initializing the 3D CNN vision backbone from the previous stage. The LM backbone is initialized with a pre-trained LLaMA-2-7B. We train the model for $2$ epochs using a learning rate of $5\times10^{-6}$. We use a batch size of $32$, gradient norm clipping of 1.0, and the AdamW optimizer \cite{loshchilov2017decoupled} with betas $0.9$ and $0.95$ with a weight decay of $0.01$. The loss objective is a standard cross-entropy loss for next-token prediction.

\myparagraph{Data Preparation.} In this stage, we fine-tune using the fitness feedbacks and fitness questions annotations on the \qevdshortclipdataset{} collection. Sequences are prefixed with the system message below, followed by a sequence of \verb|<|next\verb|>| tokens equal to the video features length, and finally the question-answer pair from the dataset. The loss on all but the answer tokens and \verb|<|next\verb|>| are $0$ in this stage.

\begin{quote}
    \scriptsize
   \verb|<|system\verb|>|You are an expert fitness coaching AI who coaches users as they exercise. You observe them silently, assess their performance, 
   and answer any questions they have.\verb|</|system\verb|>|
\end{quote}

For short video clips, we always use the request ``Please provide a feedback for the user.'' in place of the question.

\subsection{Long-range Videos}
LoRA~\cite{hu2021lora} is used for fine-tuning our model in this stage on \qevdfulldataset{} with a learning rate of $1\times10^{-6}$ and LoRA dimension as $32$. Other training details remain the same as the previous stage.

\myparagraph{Data preparation.} Due to memory constraints, we train on individual exercise segments which corresponds to a length of roughly $30$ seconds. We leave it to future work to train across timed exercise transitions. Sequences are prepared as an interleaved sequence of \verb|<|next\verb|>| tokens and feedback response tokens. Feedback response tokens are prefixed with the special \verb|<|feedback\verb|>| action token and added in the sequence according to its ground truth timestamp. 
Finally, the sequence is prefixed with the following system prompt:

\begin{quote}
    \scriptsize
    \verb|<|system\verb|>|You are an expert fitness coaching AI who coaches users as they exercise. You assess their performance, 
    and proactively provide feedback.\verb|</|system\verb|>|
\end{quote}

\section{Baselines and Evaluation}

\subsection{LLM-Accuracy Prompt} We used the following prompt to evaluate the generated feedback:

\begin{quote}
    \scriptsize
    \verb|<|INST\verb|>|
    You are an intelligent chatbot designed for evaluating feedback sequences provided by a virtual fitness coach to a person. You always provide your responses as a python dictionary string. \\
    
    Your task is to compare the accuracy of the the predicted feedback with the ground truth feedback. Here is how you can accomplish this:
    
    -The predicted feedback must be factually accurate, relevant and align with the ground truth feedback.
    
    -Consider synonyms or paraphrases as valid matches.
    
    -Take into account repetition counts that can expressed both in numeric form or in words.\\

    Please evaluate the following predicted feedback:
    
    -Ground truth feedback: $\cdots$
    
    -Predicted feedback: $\cdots$ \\
    
    Provide your evaluation as a python dictionary string with the accuracy score where the score is an integer value between 1 and 5, with 5 indicating the highest level of accuracy. Generate the response only in the form of a Python dictionary string with keys 'score', where its value is the accuracy score in INTEGER, not STRING. DO NOT PROVIDE ANY OTHER OUTPUT TEXT OR EXPLANATION.
    For example, your response should look like this: {``score'': 3.2}.
    \verb|</|INST\verb|>|
\end{quote}

\subsection{Language-only Socratic Baseline}
    The following prompt was used to generate feedback with the socratic baseline shown in~\cref{tab:interactive_eval}:

    \begin{quote}
    \scriptsize
        You are an expert fitness coaching AI who coaches users as they exercise. You assess their performance, count repetitions, and proactively provide feedback. The user should be doing $\cdots$\\
        
        Provide a SHORT ONE SENTENCE RESPONSE to the user based on based on the activity from the last $5$ seconds shown below. Take into account what you said before but DO NOT repeat it exactly. The response should be in second-person perspective. Ask them to correct any mistakes you see otherwise provide encouraging feedback.\\
        
        For example:\\
        User activity:\\
        Timestep: 0.55 -- The user is doing the exercise.\\
        Timestep: 1.55 -- The user is doing the exercise.\
        Timestep: 2.45 -- The user was fast.\\
        Timestep: 3.45 -- The user was fast.\\
        Timestep: 4.62 -- The user was fast.\\
        Timestep: 5.42 -- The user has good form.\\
        
        Response: Woah, great speed there!\\

        Previous user activity:\\
        $\cdots$\\

        This is what you've said so far for the previous activity:\\
        $\cdots$\\

        Latest user activity:\\
        $\cdots$\\

        Response: $\cdots$\\
    \end{quote}

    Time-stamped user activity for the most recent $5$ second window are included following "Latest user activity:" using the format shown in the example. The full history of previous activity and generated feedbacks are also provided in the prompt.

\subsection{Vison-language Baselines}

The following prompt is used to generate feedback with the vision-language baseline models presented in~\cref{tab:interactive_zero_shot_eval}.

\begin{quote}
    \scriptsize
    You are an expert fitness coaching AI who coaches users as they exercise. You assess their performance, count repetitions, and proactively provide feedback. The user should be doing $\cdots$\\
    
    This is what you've said so far over the last X seconds:\\
    $\cdots$\\
    
    Provide a SHORT ONE SENTENCE RESPONSE to the user based on what you see in the video. Take into account what you said before but DO NOT repeat it exactly. The response should be in second-person perspective. Ask them to correct any mistakes you see otherwise provide encouraging feedback.\\
\end{quote}

\myparagraph{Number of frames.} The number of frames provided to the model vary across the baselines. For InstructBLIP, the latest frame is shown. For Video-LLaVA and Video-LLaMA, $8$ uniformly sampled frames from the most recent $5$ second window were shown. For Video-ChatGPT, LLaVA-NeXT-Video, and LLaVA-Vid, $20$, $8$, and $20$, frames per $5$ second interval were provided, respectively -- frames were accumulated during evaluation for each feedback interval (\textit{E.g.}, when generating a feedback at $10$s, the frames from the first $5$s interval are kept). The history of responses is provided in the prompt for all models.

\subsection{Alternative Evaluations}

To ensure the accuracy of the LLM-based auto-evaluation method described in~\cref{sec:eval_metrics}, we perform the evaluation with alternative LLMs and compare them to human evaluations. Human evaluations are done on a random subset of $200$ feedbacks. As shown in~\cref{tab:human_eval_ft}, the overall ranking order is preserved across all evaluation methods. 

\begin{table*}[!htp]
\small
\centering
\caption{Evaluation of models fine-tuned with the \fulldataset{} dataset on the \fulldataset{} benchmark. ($^\dagger$indicates results of non-interactive models evaluated at regular intervals, $^*$indicates human evaluation is conducted on a smaller set of $200$ feedbacks.)}
\label{tab:human_eval_ft}
\begin{tabularx}{\linewidth}{@{}X|c@{\hspace{0.75cm}}c@{\hspace{0.75cm}}|c|@{\hspace{0.75cm}}c@{\hspace{0.75cm}}c}
\toprule
Evaluation Method & \rotatebox{90}{Socratic-LLaMA-2-7B$^\dagger$} & \rotatebox{90}{\makecell{Video-ChatGPT \cite{abs-2306-05424}\\(fine-tuned)$^\dagger$}} & \rotatebox{90}{\model{}} & \rotatebox{90}{\makecell{\model{} \\ (w/o 3D CNN)}} & \rotatebox{90}{\makecell{\model{} \\ (w/o Action-Tokens)$^\dagger$}}\\
\midrule
Human$^*$ & 2.63 & 2.59 & \textbf{2.80} & 2.51 & 2.71\\
\midrule
 Mixtral-Instruct-0.1 \cite{abs-2401-04088} & 2.39	& 2.42	& \textbf{2.56} &	2.17 & 2.56\\
 LLaMA-3-8B-Instruct \cite{llama3modelcard} & 1.74 & 1.82 & \textbf{1.90} &	1.62 & 1.89 \\
 LLaMA-3-70B-Instruct \cite{llama3modelcard} & 2.17	& 2.33 & \textbf{2.45} &	2.11 & 2.41\\
\bottomrule
\end{tabularx}
\end{table*}

\end{document}